# COMPRESSED SENSING, ASBSR-METHOD OF IMAGE SAMPLING AND RECONSTRUCTION, AND THE PROBLEM OF DIGITAL IMAGE ACQUISITION WITH THE LOWEST POSSIBLE SAMPLING RATE

*L. P. Yaroslavsky*[*]

Dept. of Physical Electronics, School of Electrical Engineering,
Tel Aviv University, Tel Aviv, Israel

## ABSTRACT

The problem of minimization of the number of measurements needed for digital image acquisition and reconstruction with a given accuracy is addressed. Basics of the sampling theory are outlined to show that the lower bound of signal sampling rate sufficient for signal reconstruction with a given accuracy is equal to the spectrum sparsity of the signal sparse approximation that has this accuracy. It is revealed that the compressed sensing approach, which was advanced as a solution to the sampling rate minimization problem, is far from reaching the sampling rate theoretical minimum. Potentials and limitations of compressed sensing are demystified using a simple and intuitive model, A method of image Arbitrary Sampling and Bounded Spectrum Reconstruction (ASBSR-method) is described that allows to draw near the image sampling rate theoretical minimum. Presented and discussed are also results of experimental verification of the ASBSR-method and its possible applicability extensions to solving various underdetermined inverse problems such as color image demosaicing, image in-painting, image reconstruction from their sparsely sampled or decimated projections, image reconstruction from the modulus of its Fourier spectrum, and image reconstruction from its sparse samples in Fourier domain.

---

[*] E-mail: yaro@eng.tau.ac.il.

## 1. INTRODUCTION

Sampling is the primary operation in acquiring digital images. Contemporary digital display devices and image processing software imply by default that sampling over regular square sampling grids is used for image discretization. As it is well known, digital images acquired in this way are, as a rule, highly compressible. Hence, images are compressed for storage and transmission and then are reconstructed to the standard sampled representation for displaying and/or processing.

The phenomenon of ubiquitous compressibility of images raises very natural questions: what are fundamental reasons of the compressibility of images sampled in the conventional way and is it possible just directly measure the minimal amount of data that won't end up being thrown away? These questions were apparently first posed by the inventors of the compressed sensing approach as a solution to this problem [1, 2, 3]. Since its introduction, the compressed sensing approach to digital image acquisition has gained a considerable popularity. It has also begotten many expectations of revolutionizing the sensor industry because it is widely believed that compressed sensing can allow high-resolution information to be recovered from low-resolution measurements made with relatively smaller detector arrays and that compressed sensing allows image sampling with sub-Nyquist sampling rates [4-8].

Potentials of compressed sensing are widely advertised in the literature. Much less is known about its limitations. Particularly important is the question: how close is the amount of measurements required by compressed sensing for image acquisition to the theoretical minimum defined by the sampling theory? The present paper discusses potentials and limitations of the compressed sensing approach from the view point of the sampling theory, shows that it requires substantially more measurements than the minimum defined by the sampling theory, discusses other limitations of compressed sensing, and describes an alternative image sampling and reconstruction method that allows to draw near the theoretical minimum of the image sampling rate.

The paper is arranged as following. In Sect. 2 basics of the modern sampling theory are outlined. In Sect.3 sampling redundacy of compressed sensing is discussed and a simple physical interpretation of compressed sensing limited capability of reconstructing signals sampled with aliasing is offered. In Sect. 4 an Arbitrary Sampling and Bounded Spectrum Reconstruction (ASBSR-) method is described that enables sampling images with rates close to the theoretical minimum. In Sect. 5 results of experimental verification of the method are provided. In Sect. 6 some practical issues of application of the ASBSR-method are discussed. In Sect. 7 the possibilities of applying the ASBSR-method to solving various other underdetermined inverse problems are discussed. The concluding Sect.8 summarizes the results.

# 2. WHAT IS THE MINIMAL NUMBER OF SAMPLES REQUIRED FOR IMAGE RECONSTRUCTION WITH A GIVEN ACCURACY?

Signal sampling is based on the idea of signal band-limited approximation. For 1D signals, sampling is quite simple. Here is a signal sampling and reconstruction protocol:

- for a signal $s(x)$ to be sampled in its coordinate $x$, define an admissible Mean Square Error (MSE) $\sigma^2$ of its approximation;

- assuming that signal Fourier spectrum $\Im(f) = \int_{-\infty}^{\infty} s(x)\exp(i2\pi f x)dx$ is concentrated around the signal DC-component at $(f = 0)$, determine a frequency interval $[-B \le f \le B]$ that contains $(E - \sigma^2)/E$-th fraction of signal energy $E = \int_{-\infty}^{\infty} |\Im(f)|^2 df$;

- pass the signal through the ideal sampling low-pass filter *SLPF* with frequency bandwidth $[-B, B]$ to obtain its band-limited approximation $\tilde{s}(x)$:

$$\tilde{s}(x) = \int_{-\infty}^{\infty} s(\xi) SLPF(x - \xi) d\xi = \int_{-B}^{B} \Im(f) \exp(-i2\pi x f) df \quad (1)$$

and sample this signal with sampling interval $\Delta x = 1/2B$ to obtain signal samples:

$$\tilde{s}_k = \tilde{s}(k\Delta x) = \int_{-\infty}^{\infty} s(\xi) SLPF(k\Delta x - \xi) d\xi, \quad k = ...., -2, -1, 0, 1, 2, ..... ; \quad (2)$$

- For signal reconstruction, pass a comb signal $\left[\sum_{k=-\infty}^{\infty} \tilde{s}_k \delta(\xi - k\Delta x)\right]$, where $\delta(.)$ is the Dirac delta-function, through the ideal reconstruction low-pass filter *RLPF* with frequency bandwidth

$[-B, B]$ to reconstruct the band-limited approximation $\tilde{s}(x)$ of the signal $s(x)$:

$$\tilde{s}(x) = \sum_{k=-\infty}^{\infty} \tilde{s}_k RLPF(x - k\Delta x). \tag{3}$$

Point spreads functions **SLPF** and **RLPF** of sampling and reconstruction ideal low pass filters are linked by the relationship **$RLPF = \Delta x SLPF$**.

If a signal has a carrier frequency, say, $f_0$, i.e., if its Fourier spectrum is concentrated around frequency $f = \pm f_0 \neq 0$, the signal should first be demodulated by a signal $\sin(2\pi f_0 x)$ to make its spectrum concentrated around frequency $f = 0$. Then the demodulated signal can be sampled as it is described above. For reconstruction of the signal from its samples, one should first reconstruct the demodulated signal and then modulate the reconstruction result with the same modulating signal $\sin(2\pi f_0 x)$. Note that the sampling rate remains the same $2B$ no matter what is the carrier frequency $f_0$.

Signal sampling rate $2B$ is frequently called the "Nyquist frequency" referring to Harry Nyquist, who in 1920-th made a guess that the sufficient for signal reconstruction number of signal samples per unit time is twice the signal bandwidth. Later V. A. Kotelnikov ([9]) and C. Shannon ([10]) formulated and proved this fact as a mathematical theorem, the Kotelnikov-Shannon sampling theorem.

The Kotelnikov-Shannon sampling theorem refers to band-limited signals. In reality, signals are never band-limited. For non band-limited signals, the sampling theory states that the signal sampling rate $2B$ is the minimal sampling rate, which secures signal reconstruction with MSE equal to the signal energy outside the bandwidth $[-B, B]$. If signals are not pre-filtered before sampling by the ideal low pass filter and/or the signal reconstruction filter is not the ideal low-pass filter, which is the case in reality, aliasing artifacts appear. This why in reality signal reconstruction MSE is larger than the above minimal one.

For sampling images as 2D signals, a protocol similar to that for 1D sinals is customarily used:

- for a given image $s(x,y)$ to be sampled in coordinates $(x,y)$, choose sampling intervals $(\Delta x, \Delta y)$ (or sampling rate, in "dots per inch") over a regular rectangular sampling grid;
- pass the image through the 2D ideal sampling low pass filter with frequency bandwidth $(-1/2\Delta x, 1/2\Delta x; -1/2\Delta y, 1/2\Delta y)$ and sample the obtained image band limited approximation $\tilde{s}(x,y)$ with the chosen sampling intervals to obtain image samples

$$\tilde{s}_{k,l} = \int_{-\infty}^{\infty}\int_{-\infty}^{\infty} s(x,y) SLPF(k\Delta x - x, l\Delta y - y) dx dy =$$

$$\int_{-1/2\Delta x}^{1/2\Delta x}\int_{-1/2\Delta y}^{1/2\Delta y} \Im(f_x, f_y) \exp[-i2\pi(k\Delta x f_x + l\Delta y f_y)] df_x df_y$$

$$k = ...., -2, -1, 0, 1, 2, ....., l = ...., -2, -1, 0, 1, 2, ....., \qquad (4)$$

where $\Im_s(f_x, f_y) = \int_{-\infty}^{\infty}\int_{-\infty}^{\infty} s(x,y) \exp[i2\pi(f_x x + f_y y)] dx dy$ is image Fourier spectrum and $(f_x, f_y)$ are image spatial frequencies.

For image reconstruction from its samples $\{\tilde{s}_{k,l}\}$, pass a 2D comb signal $\left[\sum_{k=-\infty}^{\infty} \tilde{s}_{k,l} \delta(x - k\Delta x) \delta(y - l\Delta y)\right]$ through the ideal reconstruction low-pass filter $RLPF$ with frequency bandwidth $(-1/2\Delta x, 1/2\Delta x; -1/2\Delta y, 1/2\Delta y)$ to produce a band-limited approximation $\tilde{s}(x,y)$ of the image $s(x,y)$:

$$\tilde{s}(x,y) = \sum_{k=-\infty}^{\infty}\sum_{l=-\infty}^{\infty} \tilde{s}_{k,l} RLPF(x - k\Delta x, y - l\Delta y) \qquad (5)$$

with approximation MSE defined by the image energy outside the band-limiting rectangle $(-1/2\Delta x, 1/2\Delta x; -1/2\Delta y, 1/2\Delta y)$ called the sampling baseband:

$$\sigma^2 = \int_{-\infty}^{\infty}\int_{-\infty}^{\infty} |\Im_s(f_x, f_y)|^2 df_x df_y - \int_{-1/2\Delta x}^{1/2\Delta x}\int_{-1/2\Delta y}^{1/2\Delta y} |\Im_s(f_x, f_y)|^2 df_x df_y =$$

$$E - \int_{-1/2\Delta x}^{1/2\Delta x}\int_{-1/2\Delta y}^{1/2\Delta y} |\Im_s(f_x, f_y)|^2 df_x df_y \,, \tag{6}$$

This, in fact, is how commonly used image sampling and display devices work. The role of sampling and, correspondingly, image reconstruction low-pass filters is, as a rule, played by optics of image sensors in combination with image sensor and image display device apertures. Sampling intervals $(\Delta x, \Delta y)$ are chosen usually on the base of knowledge of resolving power of imaging optics (in "lines per mm") or of visual assessment of images, assuming that one needs at least, say, 2 samples for reproducing the sharpest edges in the image.

As it was already mentioned, all image display devices and image processing software imply by default that images are sampled over square sampling grids with sampling intervals $\Delta x = \Delta y$. With such a sampling, the number $N$ of image samples is equal to the image area $S_{xy}$ times the area of the sampling baseband $S_\Omega = 1/\Delta x^2$: $N = S_{xy}/\Delta x^2 = S_{xy} S_\Omega$.

If all the image Fourier spectrum largest components that contain $(E - \sigma^2)/E$-th fraction of signal energy $E$ perfectly fill a square shape of the sampling baseband $(-1/2\Delta x, 1/2\Delta x; -1/2\Delta x, 1/2\Delta x)$, $N = S_{xy}/\Delta x^2$ will be the minimal number of samples sufficient for image reconstruction with MSE $\sigma^2$. However, in reality image Fourier spectra almost never fill square shapes. This fact is illustrated in Figure 1, where 10 test images of different kind and estimates of their Fourier spectra are shown.

Fourier spectra of the images were estimated using Discrete Fourier Transform and applying to images, before spectral analysis, a circular apodization mask in order to smoothly bring a sampled image down to zero at the edges of the sampled region and in this way to avoid as much as possible spectrum estimation errors due to boundary effects. Highlighted in the figures of image spectra are zones of spectral components that contain the largest image spectra components sufficient for image reconstruction with the same MSE as that of their JPEG encoding. We will call them "Energy compaction" (EC-) zones. One can see in these figures that EC-zones of all images occupy only a fraction of the area of the sampling baseband. This fraction, i.e., the ratio of the area of the image EC-zone to the area of the baseband, is called image spectrum sparsity.

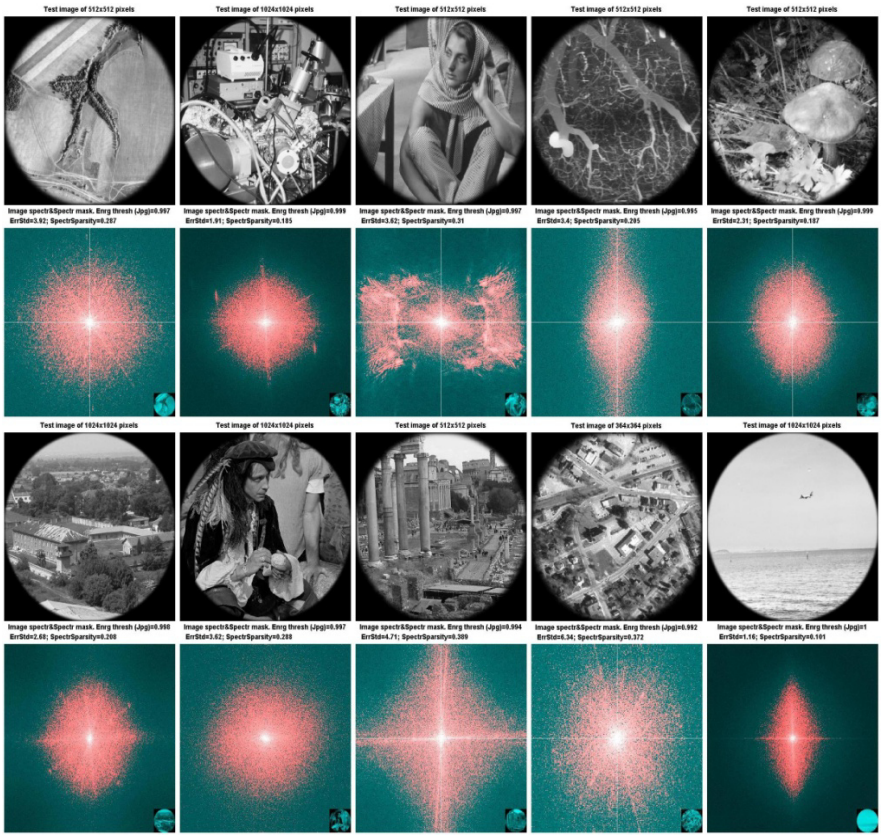

Figure 1. A set of sampled test images (first and third rows from top) and their corresponding Fourier spectra centered at spectra DC components (second and fourth rows): "AerialPhoto512", "AFM512, "Barbara512", "BloodVessels512", "Mushrooms512", "Nish1024", "Pirat1024, "Rome512", "WestConcord364" and "Test4CS1024". Highlighted are the largest spectral components sufficient for image reconstruction with the same MSE as that of their JPEG encoding.

As it was already mentioned, the minimal sampling rate of 1D signals sufficient for signal reconstruction with a given MSE $\sigma^2$ is equal to the size $2B$ of signal spectral interval that contains $(E-\sigma^2)/E$ fraction of signal energy, i.e., to the spectrum sparsity. The same is true for images as 2D signals. In order to prove that, consider the following discrete model.

Let $N$ be the number of image samples chosen for image sampled representation over a standard regular square sampling grid. Assume that available are $K < N$ image samples and it is required to reconstruct the entire image of $N$ samples with minimal MSE. Choose an image transform $\Phi_N$ and use available $K$ image samples to compute $K$ transform coefficients. Then a "Bounded Spectrum" (BS-) approximation to the entire image can be reconstructed by the inverse transform of the set of found $K$ transform coefficients supplemented with the rest of $N-K$ coefficients set to zero. Reconstruction MSE will be equal to the energy of $N-K$ coefficients set to zero. For a given transform, this error can be minimized if $K$ the largest transform coefficients are used for the reconstruction. In order to further minimize image reconstruction error, one should choose a transform with a better capability of energy compaction into the smallest number of transform coefficients. The said is the meaning of the discrete sampling theorem ([14], [15]).

In terms of this discrete model, $K$ transform coefficients that contain $(E-\sigma^2)/E$ fraction of image energy and reconstruct the image with MSE $\sigma^2$ form spectrum EC-zone of the image. The ratio $S_\sigma = K/N$ is the image spectrum sparsity in the chosen basis on this level of MSE. From the above reasoning it follows that the sampling rate (the number of samples per unit of image area) sufficient for image reconstruction with a given MSE is equal to the image spectrum sparsity on this level of MSE.

In the limit, when $N \to \infty$, the discrete model converts to a continuous one. In particular, if Discrete Fourier transform (DFT) is chosen as the image transform, it converts to the integral Fourier transform, and the discrete sampling theorem converts to the classic sampling theorem, which states that the minimal density of image samples sufficient for reconstructing images from their samples with MSE $\sigma^2$ is equal to the area $B_\Omega$ of the EC-zone $\Omega$ in the Fourier domain that contains $(E-\sigma^2)/E$-th fraction of the image signal energy $E$. If image is sampled over a square sampling grid with sampling intervals $\Delta x$, the square baseband $(-1/2\Delta x, 1/2\Delta x; -1/2\Delta x, 1/2\Delta x)$ must encompass the image spectrum EC-zone. Therefore, the number of image samples on a square sampling grid $N = S_{xy}/\Delta x^2$ will always inevitably exceed the minimal required number of samples $N_{min} = S_{xy} B_\Omega$. This is the reason for the ubiquitous compressibility of digital images. Relative, with respect to the square baseband, areas of EC-zones of test images shown in Figure 1, i.e., sparsities of their DFT spectra, range between 0.1 and 0.39, which means that the images are from 2.5 to 10 times oversampled.

## 3. COMPRESSED SENSING: WHY AND WHEN IS IT POSSIBLE TO PRECISELY RECONSTRUCT SIGNALS SAMPLED WITH ALIASING?

Compressed sensing approach is based on the idea of signal sparse approximation, i.e., approximation of signals by signals whose spectrum in a certain chosen "sparsifying" transform contains less non-zero transform coefficients than the total number of the coefficients. It was proven in the theory of compressed sensing that if an image of $N$ samples is known to have, in the domain of a certain transform, only $K$ non-zero transform coefficients out of $N$, the image can be precisely reconstructed from $M > K$ measurements by means of minimization of **L0** or, which is more practical, **L1** norm in the image transform domain, provided the following inequality holds ([11]).

$$M/K > -2\log[(M/K)(K/N)] \tag{7}$$

As it was shown in the previous section, signal sparsity $Ss = K/N$ is the theoretical minimum of the sampling rate required for signal reconstruction. Therefore, the ratio $R = M/K$ of the number of required measurements $M$ to the number $K$ of signal non-zero transform coefficients represents sampling redundancy with respect to the theoretical minimum. Inequality (7) can be rewritten as a relationship between signal sparsity $Ss = K/N$ and sampling redundancy $R = M/K$ as

$$R > -2\log(R \times Ss) \qquad (8)$$

Numerical evaluation of the relationship (8) between sampling redundancy $R = M/K$ and signal sparsity $Ss = K/N$ is presented in Figure 2 by dash-dot line. Solid line in this figure is plotted on the base of experimental data reported in the literature ([12]).

One can see in Figure 2 that, in the range of sparsities of test images shown in Figure 1 from 0.1 to 0.39, sampling redundancy of compressed sensing should theoretically be larger than 2 – 3 and, according to the experimental curve, it should be larger than 2.5 – 5. This means that sampling redundancy of compressed sensing for this set of test images is not much lower than that of their regular sampling (2.5 - 10) and that compressed sensing is quite far from reaching the theoretical minimum of signal sampling rates.

The mechanism of image acquisition and reconstruction by means of the compressed sensing approach and the nature of its sampling efficiency limitation given by Eq. 8 are in the current literature, as a rule, hidden in thickets of mathematical theorems. They can be demystified on the following simple example.

Let a signal of $N$ samples composed of a certain known number $K < N$ of sinusoidal components be subsampled arbitrarily in $M < N$ points and it is required to precisely reconstruct the signal from these $M$ samples, i.e., to determine amplitudes and frequencies of the signal sinusoidal components. Figure 3 illustrates how and when this can be done. Plots in this figure are numbered from top to bottom.

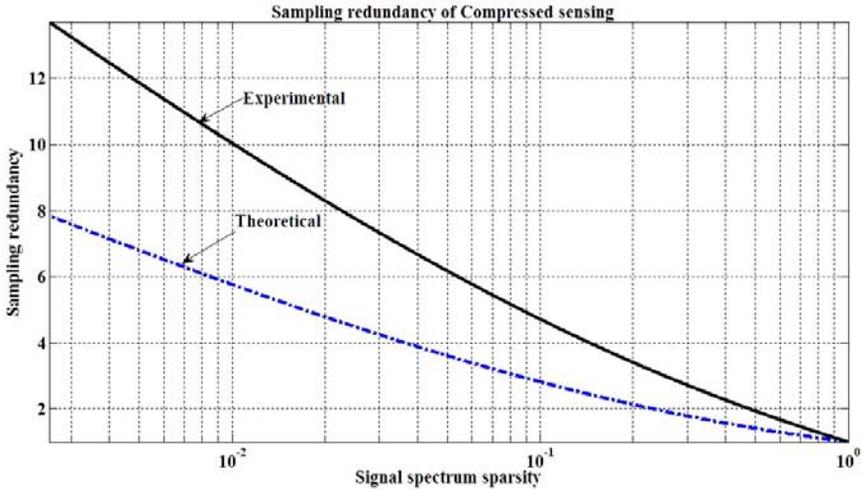

Figure 2. Theoretical and experimental relationships between signal sparsity and sampling redundancy required by the compressed sensing approach. For sampling redundancies above the curves signal reconstruction is possible, otherwise it is not.

A test signal shown in Figure 3, the 1$^{st}$ plot, is composed of three sinusoidal components ($K = 3, N = 256$) seen as three Kronecker deltas in the signal Discrete Cosine Transform (DCT) spectrum (the 2$^{nd}$ plot in Figure 3). When this signal is subsampled as it is shown in the 3$^{rd}$ plot in Figure 3, aliasing spectral components appear in the spectrum of the subsampled image. They can be seen in the 4$^{th}$ plot in Figure 3. In our example the signal is sub-sampled in random positions with sampling rate 0.15 ($M = 38$).

Intensity of the aliasing components grows with lowering the subsampling rate. Obviously, however, that, while the subsampling rate is not too low, the true spectral components of the signal exceed the highest picks of the aliasing spectral components and one can detect those true components by means of finding positions of the given number $K$ (in this particular case $K = 3$) the largest spectral components (the 4$^{th}$ plot in Figure 3). Ones these positions are determined, the following iterative Gerchberg-Papoulis type signal reconstruction algorithm can be run:

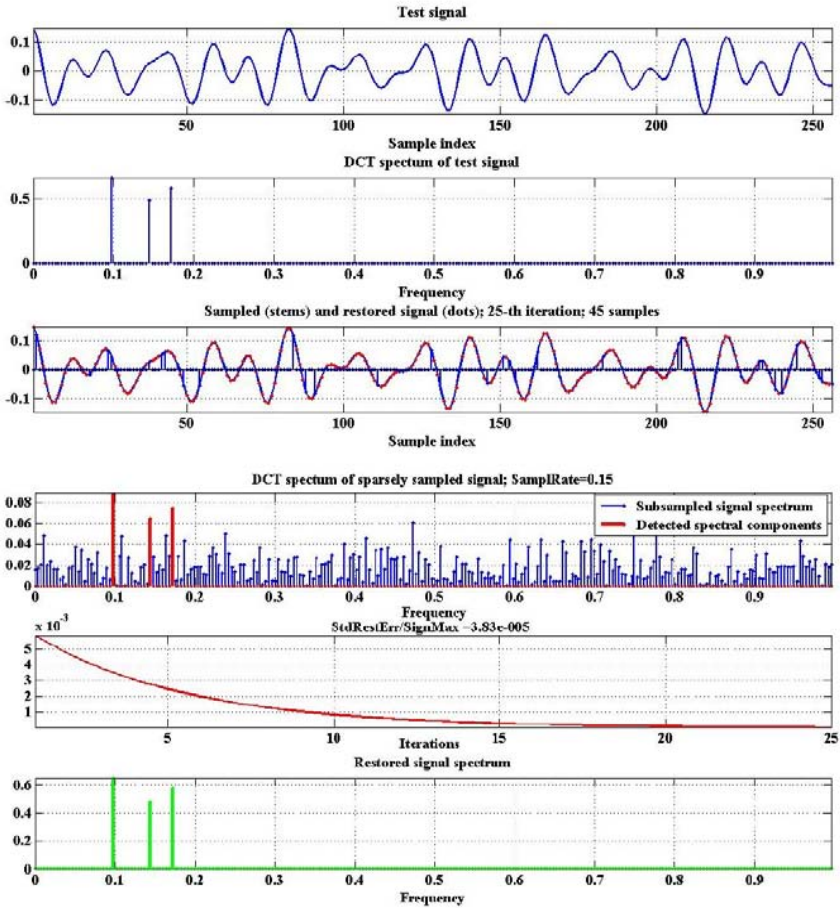

Figure 3. From top to bottom: test signal composed of 3 sinusoidal components; its DCT spectrum; this signal randomly subsampled and reconstructed; DCT spectrum of the subsampled signal; plot of reconstruction root mean square error (RMSE) vs the number of reconstruction iterations; DCT spectrum of the reconstructed signal. Frequency in plots of spectra is given in fraction of the sampling baseband.

- Compute DCT of the current estimate of the reconstructed signal.
- Detect the given number $K$ the largest spectral components.
- Set to zero all spectral components except the detected ones.
- Compute inverse DCT of the modified in this way signal spectrum to get a next estimate of the reconstructed signal.

- Replace samples of the obtained estimate of the reconstructed signal in the positions of available signal samples by the available ones.
- Repeat the loop.

The plot of reconstruction RMSE vs the number of reconstruction iterations (the 5$^{th}$ plot in ure 3) and plot of the reconstructed signal DCT spectrum (the 6$^{th}$ plot) illustrate this process and show that virtually precise reconstruction of the signal is achieved after a couple of tens of iterations: after 25 iterations reconstruction RMSE is $3.8 \times 10^{-5}$. Note that in this particular example signal sparsity is $Ss = 3/256 \approx 1.2 \times 10^{-2}$ and sampling redundancy is $R = M/K = 38/3 \cong 12.7$.

If signal subsampling rate is too low, and aliasing is severe, reliable detection of the signal spectral components in the spectrum of the sub-sampled signal and, therefore, signal reconstruction become impossible. It is clear also that the presence of noise in the sampled data hampers reliable detection of the signal spectral components and requires additional sampling redundancy for reliable signal reconstruction.

When subsampling is carried out in random positions, one can evaluate performance of the described method of signal reconstruction in terms of the probability of error in detecting signal spectral components. This probability depends on the sub-sampling rate and on the signal sparsity $K/N$.

Figure 4, left plot, presents results of experimental evaluation of the probability of frequency identification error as a function of sub-sampling rate for a sinusoidal signal ($K = 1$) of 5 different frequencies (0.9, 0.7, 0.5, 0.3, and 0.1 of the signal baseband), 5 different signal lengths $N$ (128, 256, 512, 1024, 2048), and, correspondingly, of 5 different signal sparsities $K/N$. These results for each value of signal sparsity are averages over the results obtained for different frequencies. Right plots in Figure 4 show sampling redundancies required for signal reconstruction by the above described algorithm with probability of signal frequency identification error less than $10^{-4}$, $10^{-3}$, and $10^{-2}$. The results were obtained by a Monte-

Carlo simulation of the algorithm with $5 \times 10^4$ realizations of random sampling for each individual experiment with a given sampling rate, signal frequency, and signal length.

These results illustrate the principles of the compressed sensing approach to reconstruction of signals sampled with aliasing. They clearly demonstrate why a considerable sampling redundancy is required in order to secure reliable reconstruction of such signals using a priori knowledge of the signal sparsity.

The sampling redundancy required by compressed sensing is not its only drawback. The applicability of compressed sensing is also impeded by its vulnerability to noise in sensed data and by the impossibility to predict and secure the resolving power of reconstructed images. Resolving power of images is determined by the size and shape of EC-zones of their spectra. EC-zones of spectra of images reconstructed by methods of compressed sensing are formed in the process of image reconstruction rather than are specified in advance from the requirements to image resolving power.

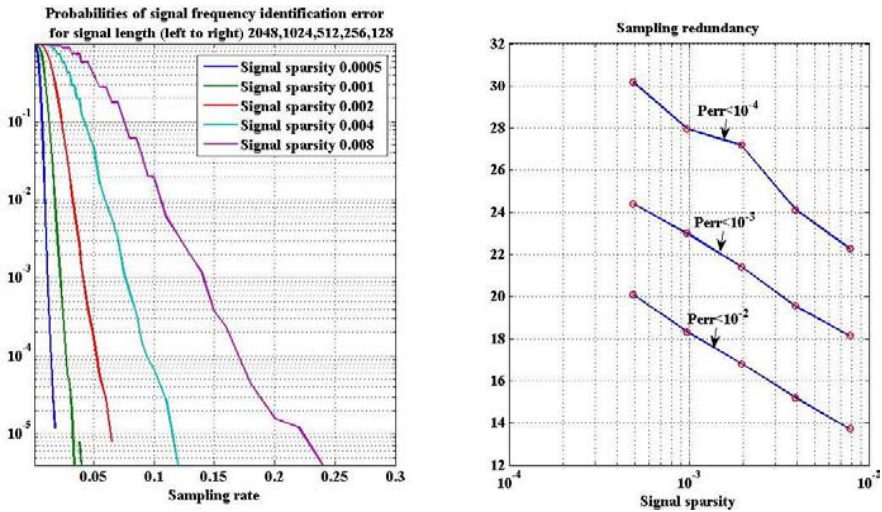

Figure 4. Plots of the probabilities of error in identification of signal frequency (left) and estimates of sampling redundancy as a function of signal sparsity $K/N$ (right), for $K = 1$ and $N = 128, 256, 512, 1024, 2048$.

To summarize, compressed sensing methods are to a certain degree capable of reconstructing sparse approximations of images sampled with aliasing. No knowledge regarding EC-zones of image spectra is required for this. One has only to choose an image sparsifying transform. Not using any knowledge regarding image spectra EC-zones has its price. Because of this, compressed sensing requires a significant redundancy in the number of sufficient for image reconstruction measurements compared to the theoretical minimum.

In many practical tasks of digital image acquisition, the assumption of the complete uncertainty regarding image spectra EC-zones has no justification. In fact, if one is ready, as it is supposed by the compressed sensing approach, to accept a sparse approximation to an image and has chosen an image sparsifying transform, one tacitly implies certain knowledge of the energy compaction capability of the chosen transform. In what follows we show that making use of this in any way available a priori knowledge allows implementing image sampling with rates close to the theoretical minimum ([13]).

## 4. ASBSR-METHOD OF IMAGE SAMPLING AND RECONSTRUCTION

As it was outlined in Sect. 2, sampling theory suggests the following principle of image sampling and reconstruction:

- Choose the required number $N$ of image samples.
- Choose an image sparsifying transform.
- Specify a desired EC-zone of the image spectrum, i.e., a set of $M \leq N$ transform coefficients to be used for image reconstruction.
- Measure $M$ image samples.
- Use $M$ image samples for determining $M$ transform coefficients of the chosen EC-zone.

- Set the rest $N - M$ transform coefficients to zero and use the obtained spectrum for reconstruction of required $N$ image samples by means of its inverse transform.

Consider possible ways for implementing this principle.

### Choosing a Transform

The choice of the transform is governed by the transform energy compaction capability. An additional requirement is the availability of a fast transform algorithm. From this point of view, Discrete Cosine Transform (DCT), Discrete Fourier Transform (DFT) and wavelet transforms are among primary candidates.

### Specification of Image EC-Zones

Specification of the subset of image transform coefficients to be used for image reconstruction, i.e., of the image spectrum EC-zone, can be made on the basis of the known capability of image transforms, such as DCT, to compact most of the image signal energy into few transform coefficients. In what follows we will assume using DCT as the image sparsifying transform. DCT is known to compact the image largest transform coefficients into more or less compacted groups in the area of coefficient's lower indices around the DC component.

Considerable practical experience, including that obtained in course of developing of zonal quantization tables for JPEG image compression standard, shows that although these groups of DCT coefficients do not have sharp borders, they are quite well concentrated. This means that the groups can be, with a reasonably good accuracy in terms of preservation of the group total energy, circumscribed by one of some standard shapes that encompass area of image low spatial frequencies and can be specified by few parameters, such as area, aspect ratio, angular orientation, etc.

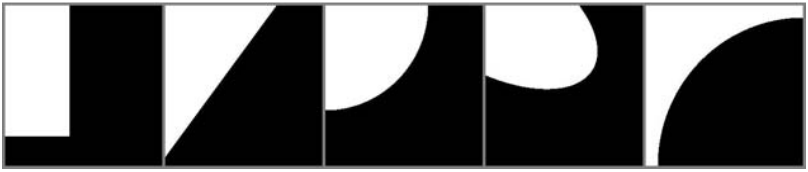

Figure 5. Examples of possible standard shapes of image EC-zones for image DCT spectra. From left to right: rectangle, triangle, pie-sector, oval, super ellipse. Spectrum DC components are in the upper left corners of the shapes.

Figure 5 presents a set of five possible standard shapes suited for DCT as the sparsifying transform: rectangle, triangle, pie-sector, ellipse, and super ellipse. In principle, each particular standard shape can be associated with a certain class of images, such as micrographs, aerial photographs, space photos, in-door and out-door scenes, etc.

The author's experimental experience shows that no fine tuning of shape parameters is required for specifying chosen shapes as approximations of image EC-zones. This property of sparse DCT spectra is illustrated in Figure 6 on an example of sparse DCT spectrum of test image "BloodVessels512". One can easily notice in the image a certain preferential prevalence of horizontal edges. This causes anisotropy of image sparse spectrum seen in boxes b) – d) in Figure 6, where shown are, marked as white dots, the largest DCT coefficients that reconstruct this image with root mean square error (RMSE) 3.85 gray levels of 255 levels, the same as the reconstruction RMSE of this image after its standard JPEG compression by the Matlab tools. These coefficients occupy 0.164 of the image baseband area and form the image spectrum EC-zone. Additionally in boxes b) – e) shown are borders of rectangular, triangular and oval shapes that all have the same area (0.275 of the baseband area) and different aspect ratios (0.35, 0.25, 0.3, and 0.45, correspondingly). When used as spectrum bounding shapes that approximate the image EC-zone, they all reconstruct the test image with practically the same reconstruction RMSEs (4.1, 3.7, 3.8, and 3.8 of image gray levels, correspondingly) as that of the JPEG compression (3.85).

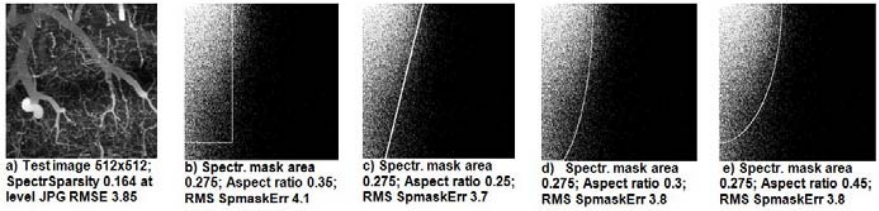

Figure 6. Test image "BloodVessels512" with spectrum sparsity 0.164 at the level of JPEG encoding RMSE 3.85 of image gray levels (a), positions of the image largest spectral components that reconstruct image with RMSE 3.85 (white dots), and borders (white lines) of approximative rectangular, triangular and oval shapes with different shape parameters ( b) –e)).

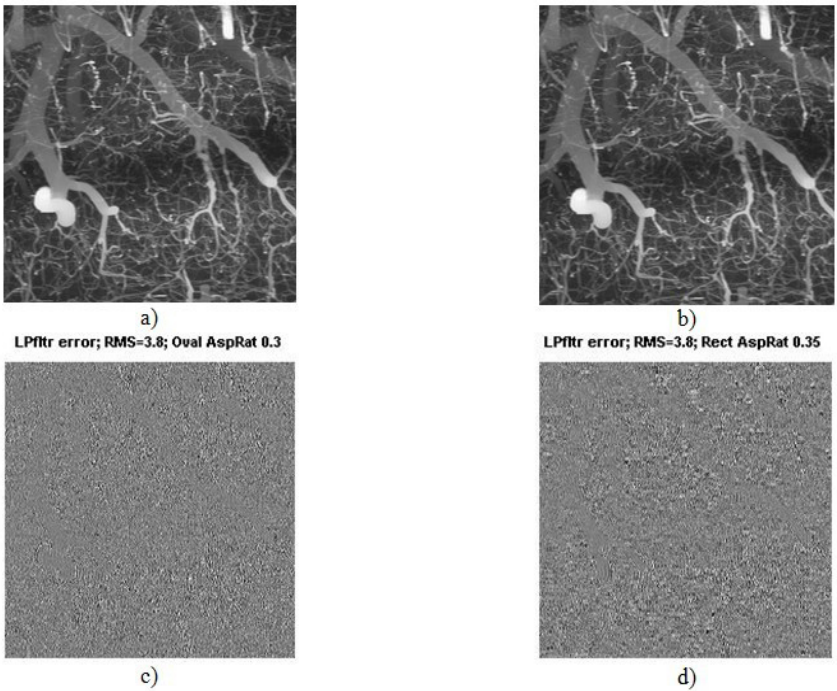

Figure 7. a), b) - Images reconstructed after bounding DCT spectrum of the test image of Figure 6, a) by oval (Figure 6, d) and by rectangular (Figure 6, b) spectral masks; c), d) - patterns of the corresponding reconstruction errors (displayed 8 times contrasted).

Images with spectra bounded by one of these shapes are are visually indistinguishable one from another, though patterns of the reconstruction errors look, naturally, a bit different. Two examples of such images obtained for spectrum bounding by the rectangle, as in Figure 6, b), and by the oval, as in Figure 6 d), are shown in Figure 7, a) and b) along with corresponding patterns of the reconstruction errors (Figure 7, c) and d)). For display purposes, reconstruction errors are shown 8 times contrasted.

As one can see in Figure 6, shapes that are chosen to approximate the image spectrum EC-zone do not include all spectral coefficients of the EC-zone. From the other side, they include some of spectrum components that do not belong to the EC-zone. Those components that do not belong to the EC-zone and happen to be inside the chosen spectral shape have, by definition, lower intensity than components of EC-zone that happen to be outside the shape. Therefore, given the energy of all spectral components encompassed by the chosen spectral shape, the number of these internal "no-EC-zone components" must exceed the number of EC-zone components not encompassed by the shape. This means that the area of the shape that defines the number of samples to be taken will always exceed to a certain degree the number of EC-zone coefficients, which, theoretically, is the minimal number of samples required. For instance, redundancies of approximating EC-zone shapes of the image in Figure 6 are $0.275/0.164 = 1.67$. Experimental results reported in the next section show that redundancies of standard EC-zones for natural images are, as a rule, of the same order of magnitude. This EC-zone shape's redundancy is the price one should pay for not knowing exact indices of transform coefficients that form the image spectrum EC-zone.

## Specifying Positioning of Image Samples

Positioning of image samples should permit computation from them transform coefficients chosen for image reconstruction. Some image transforms, such as wavelets, impose in this respect certain limitation on positions of image samples. For DFT and DCT, positions of image samples can be arbitrary ([14]) An additional advantage of using DFT and DCT as image sparsifying transforms is that they are discrete representations of the integral Fourier transform and, as such, they ideally concord with characterization of imaging systems in terms of their Modulation Transfer Functions.

## Methods of Image Reconstruction

For image reconstruction, one can consider two options:

- Direct inversion of the $M \times N$ inverse transform matrix that links $M$ available samples and $M$ transform coefficients specified by the chosen spectral EC-zone with the rest $N - M$ coefficients set to zero. Once $M$ chosen transform coefficients are found and the rest $N - M$ transform coefficients are set to zero, the inverse transform is applied to the found spectrum to reconstruct all required $N$ samples. Generally, matrix inversion is a very hard computational task and no fast matrix inversion algorithms are known. In our particular case, a pruned fast transform matrix should be inverted. There exist pruned versions of fast transforms for computing subsets of transform coefficients of signals with all samples except several ones equal to zero [15] Yaroslavsky L., *Theoretical Foundations of Digital Imaging*, CRC Press, Roca Baton,
- [16]), which is inverse to what is required in the given case. The question, whether these pruned algorithms can be inverted for computing a subset of transform non-zero coefficients from a subset of signal samples, is open.
- An iterative Papoulis-Gerchberg type algorithm. The algorithm at each iteration consists of two steps:
    (i) The iterated reconstructed image is subjected to the direct transform and then spectral coefficients outside the the chosen bounding EC-zone are zeroed for obtaining an iterated image spectrum.
    (ii) The iterated image spectrum is inverse transformed and then samples of the obtained image at positions, where they were actually taken at sampling, are replaced by the corresponding

available ones, which produces the next iterated reconstructed image. As a zero order approximation, from which the reconstruction iterations start, an image interpolated in one or another way from the available samples can be taken. A particular interpolation algorithm used in verification experiments is detailed in the next section.

The above reasoning suggests the following protocol of image sampling and reconstruction with DCT as the image sparsifying transform:

*Image sampling:*
- Choose the required image spatial resolution $SpR$ (in "dots per inch") in the same way as it is being done in the conventional image sampling.
- On the basis of evaluating the image, choose one of the standard spectral bounding shapes for bounding EC-zone of DCT spectrum and their shape parameters, such as aspect ratio for rectangle and triangle, aspect ratio and orientation angle for ellipse, etc.
- Evaluate $X$ and $Y$ dimensions $ShSzX$ and $ShSzY$ of the chosen shape using $SpR$ as the shape largest diameter.
- Specify the number of pixels $N_x \geq ShSzX$ and $N_y \geq ShSzY$ "per inch" in $X$ and $Y$ dimensions of the reconstructed image.
- For the chosen shape, evaluate the fraction $Fr$ of the area the shape occupies in the rectangle $N_x \times N_y$. This fraction times $SpR \times SpR$ determines spatial density $SpD = Fr \times SpR^2$ of samples to be taken (in "dots per square inch"). The number of samples $M$ to be taken can be then found as a product of $SpD$ and the image area $ImgSzX \times ImgSzY$: $M = SpD \times ImgSzX \times ImgSzY$.
- Choose whatever sampling grid appropriate for the available image sensor and sample the image in $M$ positions; if no other option is available, use sensor's aperture as a pre-sampling low-pass filter.

*Image reconstruction:*
- Apply to the sampled image one of the above-described reconstruction options using for specification of the image spectrum EC-zone the chosen spectrum bounding shape. In this way, an image with spectrum in the chosen transform bounded by the chosen EC-zone, or a bounded spectrum (BS-) image, will be obtained, which has the prescribed spatial resolution $SpR$.

As one can see, the described sampling protocol does not essentially differ from the ordinary standard 2D sampling protocol. The only difference is that arbitrary sampling grids can be used and evaluation of image expected spectrum shape for approximating image spectrum EC-zone is required in the suggested method in addition to the specification of the desired image resolution, which anyway is required by the standard sampling protocol.

Not much different is image reconstruction from sampled data as well. In the suggested method, low pass filtering at image reconstruction is carried out numerically by means of bounding image spectrum in the chosen transform by the chosen spectral shape. Thanks to this, the method reaches the minimal sampling rate defined by the area of the chosen spectral shape. As it was mentioned, the latter is somehow larger than the area of the image sparse spectrum, which it approximates and which defines the absolute minimum of the image sampling rate.

In view of the said the suggested image sampling and reconstruction method can be called Arbitrary Sampling and Bounded Spectrum Reconstruction (ASBSR-) method.

## 5. EXPERIMENTAL VERIFICATION OF THE METHOD

The suggested image ASBSR-method has been experimentally verified on a considerable data base of test images including ten images presented in Figure 1. In the experiments, the above-described iterative Gerchberg-

Papoulis type algorithm was used and three types of sampling grids were tested:

- "quasi-uniform" sampling grid, in which $M$ image samples are distributed uniformly with appropriate rounding off their positions to the nearest nodes of the dense square sampling grid of $N$ samples;
- uniform sampling grid with jitter, in which horizontal and vertical positions of each of $M$ samples are randomly chosen, independently in each of two image coordinates, within the primary uniform sampling intervals;
- totally pseudorandom sampling grid, in which positions of samples are uniformly distributed in a pseudo-random order over nodes of the dense sampling grid of $N$ samples.

As an image transform that compacts the image spectrum, the Discrete Cosine Transform was used. As an admissible RMSE of approximation of test images by images with sparse DCT spectra, RMSE of image compression by the standard JPEG compression in Matlab implementation is taken. These RMSEs were used for choosing areas of EC-zone shapes for each particular image. The chosen shapes were used for bounding image DCT spectra both for image pre-filtering before sampling and in the process of image reconstruction. The former is important not only for avoiding aliasing artifacts but also for securing convergence of the iterative reconstruction algorithm to an image with spectrum bounded by the chosen EC-shape.

As a strarting zero-order approximation to reconstructed images, each not available image sample was interpolated from three nearest to it available samples taken with weights inversely proportional to their distances from the interpolated sample.

Figure 8, Figure 9 and Figure 10 illustrate results of experiments with ten images of the tested set. Shown in Figure 8 are: (i) test image, (ii) reconstructed image, (iii) sampled test image, (iv) the border of the chosen shape of image EC-zone (solid line) and positions of the image DCT

spectrum largest coefficients (white dots), which reconstruct image with RMSE equal to that of image JPEG compression; (v) plots of RMS of all reconstruction errors (difference between the initial test image pre-filtered for bounding its spectrum before sampling and the reconstructed image) and of RMS of the smallest 90% of reconstruction errors vs the iteration number. Separate counting RMS of the smallest 90% of reconstruction errors is motivated by the observation that the iterative reconstruction converges not uniformly over the image area: most of the errors decay with iterations much more rapidly than few isolated large errors. Reconstruction RMSE are given in units of image gray levels in the range 0-255.

In Figure 9 and Figure 10 shown are, for the sake of saving space, only (i) reconstructed images (left column), (ii) positions of the largest DCT coefficients of sparse approximations to the corresponding test images and borders of their chosen EC-zones (middle column), and (iii) plots of reconstruction RMSE versus the number of iterations (right column).

For all shown images, sampling over uniform sampling grids with jitter was used, for which reconstruction errors decayed with iterations most rapidly. For the same number of iterations, reconstruction RMSE for totally random sampling grids was about 1.5-2 times and for "quasi-uniform" sampling grids 2-2.5 times larger than that for the "uniform with jitter" sampling grid. For "quasi-uniform" sampling grids, stagnation of the iteration process was observed, which can apparently be attributed to the presence of regular patterns of thickenings and rarefications of sampling positions due to rounding off their coordinates to positions of nodes of the regular uniform sampling grid.

Numerical results for reconstruction accuracy, spectrum sparsity, sampling rate, redundancy of the chosen EC-zones (the ratio of fractions of area they occupy in the sampling baseband to the spectrum sparsity), sampling redundancy (the ratio of the sampling rate to the relative area of the chosen EC-zone), and the overall sampling redundancy (the ratio of the sampling rate to the spectrum sparsity) obtained for all ten test images shown in Figure 8, Figure 9 and Figure 10 are presented in Table 1.

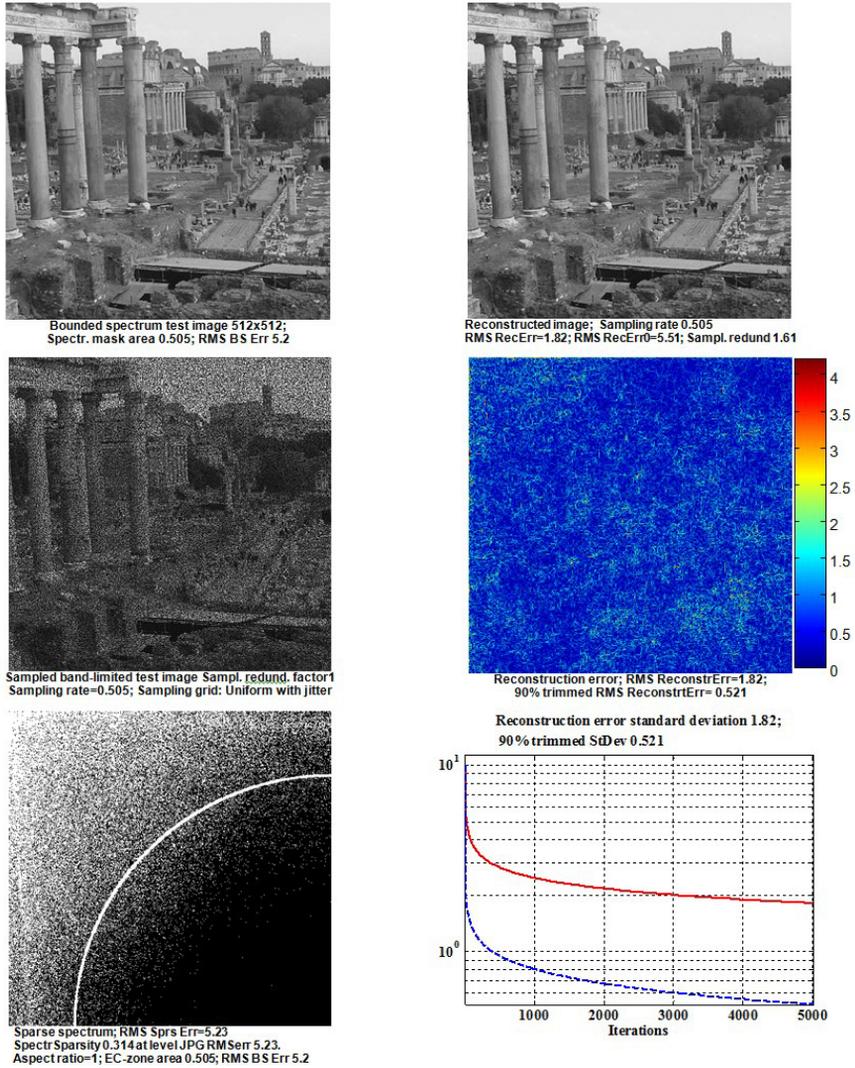

Figure 8. Results of experiments on sampling and BS-reconstruction of test image "Rome512". From top to bottom, from left to right: test image; reconstructed BS-image; sampled test image; test image sparse spectrum (white dots) and the border of the chosen EC-zone (white solid line); color coded (Matlab color map "jet") absolute value of reconstruction error (difference between initial and reconstructed images); plots of RMS of all (solid line) and of the smallest 90% (dash line) of reconstruction errors vs the number of iterations.

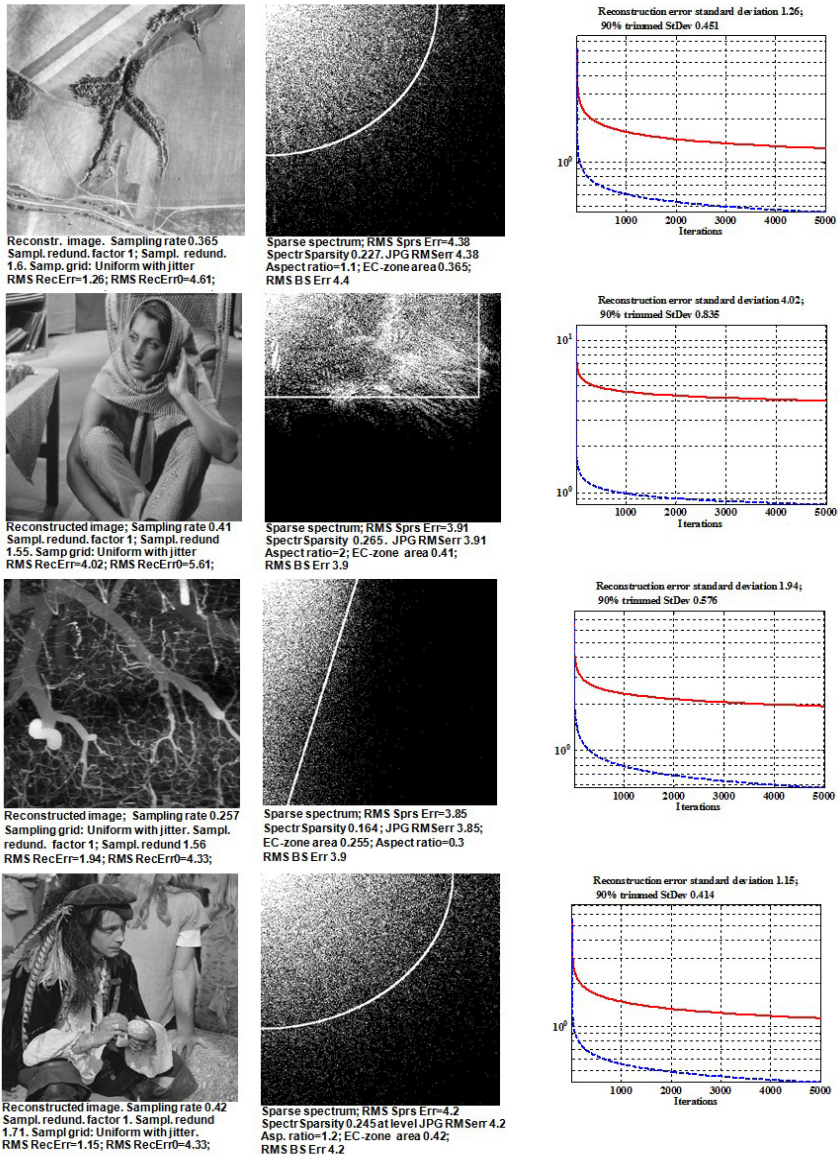

Figure 9. Results of experiments on sampling and reconstruction of test images. From top to bottom: "AerialPhoto512" "Barbara512", "BloodVessels512", "Pirat1024". From left to right: reconstructed images; image sparse spectra (white dots) and borders of the corresponding chosen EC-zones (white solid line); plots of RMS of all (solid line) and of the smallest 90% (dash line) of reconstruction errors vs the number of iterations.

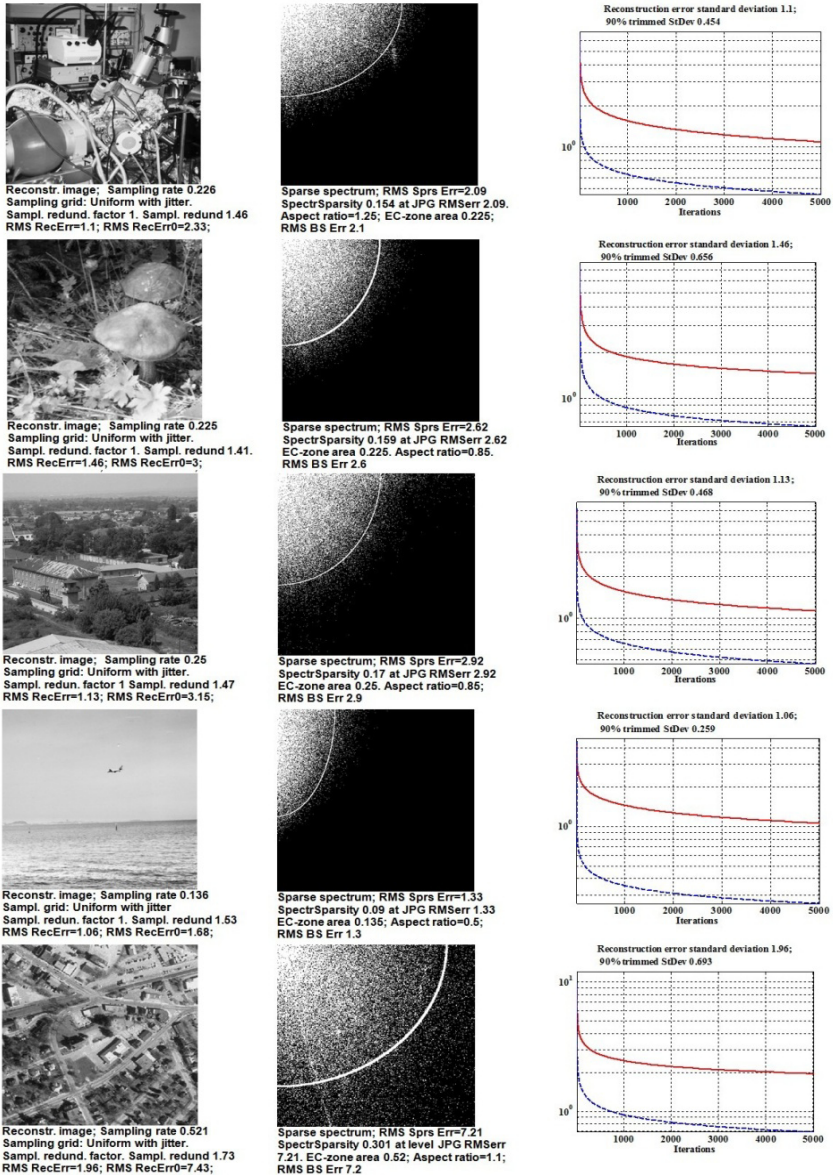

Figure 10. Results of experiments on sampling and reconstruction of test images. From top to bottom: "AFM1024, "Mushrooms512", "Nish1024", "Test4CS1024", "WestConcord364". From left to right: reconstructed images; image sparse spectra (white dots) and borders of the corresponding chosen EC-zones (white solid line); plots of RMS of all (solid line) and of the smallest 90% (dash line) of reconstruction errors vs the number of iterations.

Plots of reconstruction RMSEs vs the number of iterations in Figure 8, Figure 9 and Figure 10 show that RMS of reconstruction errors decay at the first couple of hundreds iterations quite rapidly but after they reach the value of about 2-3 quantization intervals, the error decay slows down. It was found in the experiments that one can substantially accelerate the error decay, if the number of samples is taken with a certain small redundancy, i.e., 10-20% larger than the minimal number equal to the area of the chosen EC-zone (see results for test image "Barbara512" in Table 1).

To summarize, the experiments confirm that images sampled with sampling rates equal to the minimal rate for their chosen EC-zones can be reconstructed with a sufficiently good accuracy. The redundancy in the number of required samples associated with that of standard shapes of image spectra EC-zones is of the order 1.5-1.6 and it never exceeded 2 in experiments with other images. These figures are estimates of the sampling redundancy of the suggested ASBSR-method.

**Table 1. Summary of experimental results**

| Test image | Reconstruction RMSE (Peak signal to RMSE ratio) | Spectrum sparsity | Sampling rate | Sampling redundancy ||| 
|---|---|---|---|---|---|---|
| | | | | Redundancy of chosen EC-zone | Sampling redundancy with respect to chosen EC-zone | Overall sampling redundancy |
| AerialPhoto512 | 1.26 (46.2 dB) | 0.227 | 0.365 | 1.6 | 1 | 1.6 |
| AFM1024 | 1.1 (47.3 dB) | 0.154 | 0.226 | 1.46 | 1 | 1.46 |
| Barbara512 | 4.02 (36.1 dB) | 0.265 | 0.412 | 1.55 | 1 | 1.55 |
| | 0.69 (51.4 dB) | 0.265 | 0.474 | 1.55 | 1.15 | 1.78 |
| BloodVessels512 | 1.94 (42.4 dB) | 0.164 | 0.257 | 1.56 | 1 | 1.56 |
| Mushrooms512 | 1.46 (44.9 dB) | 0.159 | 0.226 | 1.41 | 1 | 1.41 |
| Nish1024 | 1.13 (47.1 dB) | 0.17 | 0.25 | 1.47 | 1 | 1.47 |
| Pirat1024 | 1.15 (46.9 dB) | 0.245 | 0.42 | 1.61 | 1 | 1.61 |
| Rome512 | 1.92 (42.5 dB) | 0.314 | 0.5 | 1.59 | 1 | 1.59 |
| Test4CS1024 | 1.06 (47.7 dB) | 0.09 | 0.136 | 1.53 | 1 | 1.53 |
| WestConcord | 1.96 (42.3 dB) | 0.314 | 0.521 | 1.73 | 1 | 1.73 |

## 6. SOME PRACTICAL ISSUES

In this section, four issues of practical application of the suggested image ASBSR- method are addressed: (i) how robust is the method to the presence of noise in sensor data; (ii) practical considerations regarding choosing the shape of EC-zone for bounding image spectra for image sampling and reconstruction, (iii) image antialiasing pre-filtering, and (iv) computational complexity of the method.

### Noise Robustness of the ASBSR-Method

From the method description in Section 4, one can see that the ASBSR-method, just as the conventional sampling and reconstruction, is linear, i.e., it satisfies the superposition principle. No parameter of sampling and reconstruction algorithms depends on signal values and, in particular, on whether noise is present in the signal or not. Therefore, sampling and reconstruction of an image that contains additive noise will result in a reconstructed image that also contains additive noise with power spectrum bounded by the shape of the spectrum EC-zone used for image reconstruction. In particular, if the sensor noise with variance $\sigma^2$ has uniform power spectrum within the sampling baseband, noise in the reconstructed image will have variance $\kappa\sigma^2$, where $\kappa < 1$ is the relative area of the reconstruction EC-zone, and its power spectrum will be uniform within the spectrum EC-zone and zero outside it.

In order to illustrate the said, an experiment on sampling and reconstruction of an image with and without additive noise was conducted. The results are presented in Figure 11, from which one can see that the presence of noise in sampled data has no influence on the work of the reconstruction algorithm.

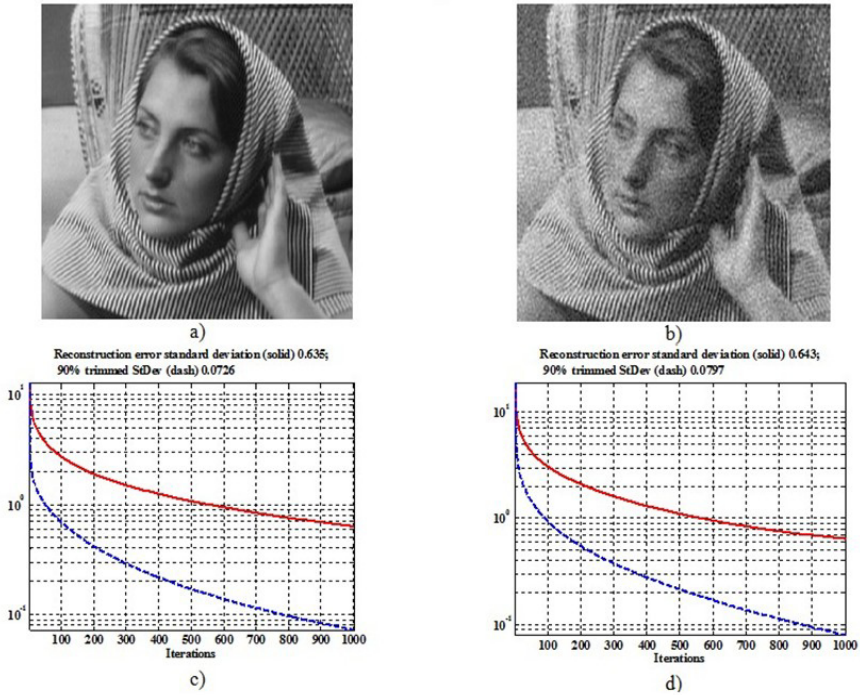

Figure 11. a),b) - images reconstructed from a sampled noiseless test image and, correspondingly, from the same sampled image contaminated with additive uncorrelated Gaussian noise with standard deviation 20 gray levels; c),d) - corresponding graphs of RMS of all (solid lines) and of the smallest 90% (dash lines) of reconstruction errors versus the number of iterations.

## Choosing the Shape of EC-Zone for Bounding Image Spectra for Image Sampling and Reconstruction

As it was mentioned in Sect. 4, no fine tuning is required for specifying shape parameters. Therefore, it is suggested that several standard shapes, such as those shown in Figure 5, should be found for different classes of images, e.g., landscape, portrait, micrographs, aerial and space photographs of different kind, and alike. This can be based, for instance, on a machine learning algorithm trained on various image data bases. For sampling images in a particular application, the user should only specify an image class. Note that specifying an image class is the standard option for setting parameters of modern digital cameras.

## Image Antialiasing Pre-Filtering

As it was indicated in the previous section, image pre-filtering before sampling is necessary in order to avoid aliasing artifacts and to secure convergence of the iterative algorithm to an image with bounded spectrum. In ordinary imaging devices, antialiasing pre-filtering is carried out by imaging optics together with apertures of image photo sensors. ASBSR-method envisages, generally, choosing an EC-zone shape for every particular image. Ordinary photo sensors are not capable of implementing this choice. As a solution to this problem, usage of multiple aperture sensors, such as those suggested in Ref. [17], can be proposed. The required effective aperture of these sensors defined by inverse Fourier transform of the chosen EC-zone shape is synthesized by combining, with appropriate weights, signals from individual apertures.

The use of conventional single aperture sensors is also possible at the expense of a certain worsening of accuracy of approximating images by their BS-copies, reconstructed by the ASBSR-method. As such a universal ("all-purpose") EC-zone shape, a "pie-sector" shape can be suggested, which fits majority of natural images quite well. As it was mentioned in Sect. 4, substantial variations of EC-zone shape parameters do not lead in practice to substantial variations of image approximation RMSE. This is why the worsening image approximation accuracy due to the use an "all-purpose" shape instead of a "dedicated" one will be, as a rule, not dramatic. For instance, total RMS error 4.72 of reconstruction of test image "BloodVessels512" using an "all-purpose" pie-sector zone (Figure 12) is only 7.7% larger than reconstruction RMSE 4.38 obtained, when a triangular EC-zone that fits image spectrum (Figure 9) was used.

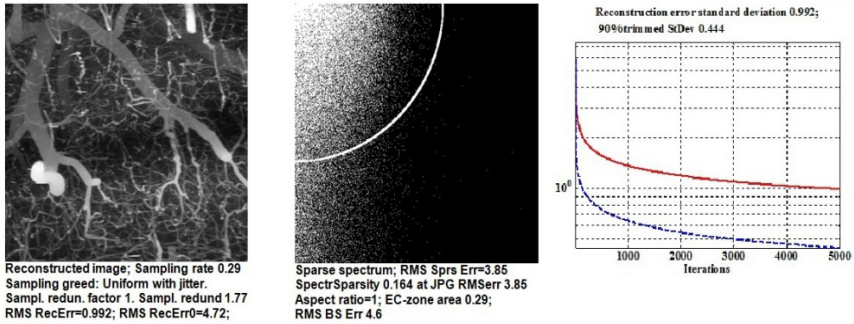

Figure 12. An example of reconstruction of test image "BloodVessels512" using a pie-sector EC-zone of the same area as that of the triangular shape (Figure 9), which better fits the image sparse spectrum.

## Computational Complexity

The computational complexity of the reconstruction algorithm per iteration is determined by the complexity $O(2N \log N)$ of floating point operations for direct and inverse fast transforms plus $O(N)$ replacement operations for sample-wise modifications of data in image domain ($M$ operations) and in its transform domain ($N - M$ operations). The order of magnitude of time required for one iteration can be estimated from these data: elapsed time for Matlab direct or inverse DCT of an array of $N = 512 \times 512$ numbers implemented on a PC "Lenovo ThinkPad X201" with processor Intel i7 and operating system Windows-7 is 52 msec.

## 7. OTHER POSSIBLE APPLICATIONS OF THE ASBSR-METHOD OF IMAGE SAMPLING AND RECONSTRUCTION

The above discussed task of reconstruction of images of $N$ samples from $M < N$ sampled data can be considered as a special case of under-determined inverse imaging problems. One can expect that the found solution to this problem, the bounded spectrum (BS-) image reconstruction, may find application for solving other under-determined inverse imaging problems as well. We illustrate this possibility in five applications: (i) demosaicing color images, (ii) in-painting of occlusions in images, (iii) image reconstruction from their sparsely sampled or decimated projections, (iv) image reconstruction from their sparsely sampled Fourier spectra, and (v) image reconstruction from the modulus of its Fourier spectrum.

**Demosaicing Color Images**

Individual cells of light sensitive arrays of modern color cameras are split into three groups allocated for corresponding Red, Green and Blue components of color images. It is made by means of placing, in front of cells, arrays of red, green and blue filters. Most frequently used is Bayer's arrangement of filters ([18]) shown in Figure 13, a). We will call this arrangement "Regular Bayer". There is also an alternative method shown in Figure 13, b). It assumes the regular Bayer arrangement of green pixels and pseudo-random arrangements of red and blue pixels ([19]). We will call this arrangement "Semi-random". In both arrangements, density of green pixels is 0.5, and densities of red and blue pixels are correspondingly 0.25 of the overall pixel density, i.e., half of all cells of the sensor array are allocated for green pixels, one quarter is allocated for red, and one quarter for blue pixels.

Images captured by color cameras are displayed in full resolution of the sensor, for which purpose half of pixels of green image components and three quarters of pixels of red and blue image components should be reconstructed from their corresponding available pixels. This reconstruction of unavailable color separated pixels is called image demosaicing. As a demosaicing mechanism, different interpolation methods are discussed in literature ( [19]). In practice, bilinear interpolation is ordinarily used. We will call this method of color image demosaicing "Bilinear demosaicing".

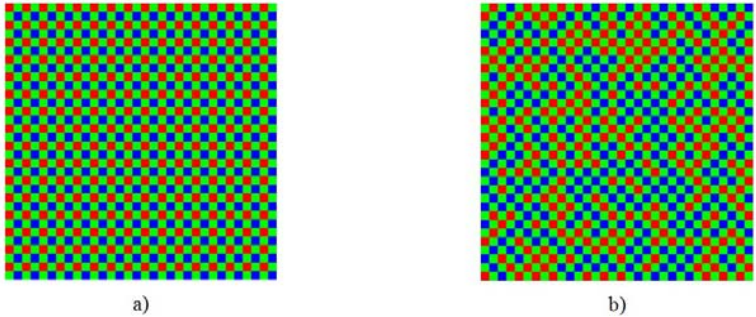

Figure 13. Regular Bayer (a) and semi-random (b) arrangements of color separated pixels in photosensitive arrays of modern color cameras.

Bilinear interpolation is one of the simplest 2D interpolation methods. It, however, is far from being perfect since it distorts images in their baseband and introduces spectrum aliasing errors. From the theory outlined in Sect. 2, it follows that, given the number of available pixels, optimal image reconstruction in terms of reconstruction RMSE is the Bounded Spectrum (BS-) reconstruction, i.e., reconstruction of images with spectrum bounded by chosen for them EC-zone shapes. In application to image demosaicing, the most natural is using as the spectrum bounding shape the "all-purpose" shape in form of a pie-sector that allows implementing image anti-aliasing pre-filtering before their sampling by camera optics and by apertures of cells of its photo sensitive arrays. We will call this method of demosaicing "BS-demosaicing" Performance of image BS-demosaicing in comparison with that of Bilinear demosaicing is illustrated in Figure 14 on a test image LightHouse512 for two cases: regular Bayer arrangement and Semi-random arrangement of pixels, correspondingly.

In the top row in Figure 14 shown are (from left to right) a test color image of 512x512 pixels and sparse spectra (white dots) of the image red, green and blue components that contain spectral coefficients, which reconstruct corresponding images with reconstruction RMSE equal to that of JPEG image coding. Solid lines in the spectra images indicate borders of EC-zones used for image reconstruction by the BS-method. According to the rates of pixels in red, green, and blue image components, relative areas of the EC-zones are set to 0.25, 0.5, and 0.25, correspondingly. In the middle and bottom rows of the figure shown are bottom right 128x128 pixel fragments of the test and of demosaiced images obtained by the BS- and by Bilinear demosaicing methods for regular Bayer (middle row) and semi-random (bottom row) color pixel arrangements. The fragments are magnified for better visibility of demosaicing artifacts. One can easily see in these images that the Bilinear demosaicing produces much more color artifacts compared to those of the BS-demosaicing, for which the artifacts are quite subtle.

Numerical data on reconstruction RMSE for both demosaicing methods and Regular Bayer and Semi-random pixel arrangements are given in Table 2. According to them, BS-demosaicing outperforms the Bilinear Bayer demosaicing in terms of reconstruction RMSE as well. In particular, total reconstruction RMSE for the BS- and Bilinear methods are, correspondingly, 7.98 and 9.76 for the case of "Regular Bayes" arrangement and 8.23 and 11.05 for "Semi-random" arrangement of color pixels. Relatively large values of image reconstruction errors can be attributed to the fact that, as one can see from image spectra, quite large portions of spectra red, green, and, especially, of blue components fall outside the EC-zones used for image reconstruction.

**Table 2. RMS reconstruction errors for the Bilinear and BS-demosaicing**

| "Regular Bayes" color pixel arrangement | | | | |
|---|---|---|---|---|
| BS-demosaicing, RMS of reconstruction errors | Red component 9.11 | Green component 4.38 | Blue component 9.43 | Total average 7.98 |
| Bilinear demosaicing RMS of reconstruction errors | Red component 11 | Green component 6.31 | Blue component 11.2 | Total average 9.76 |
| "Semi-random" color pixel arrangement | | | | |
| BS-demosaicing, RMS of reconstruction errors | Red component 9.55 | Green component 4.36 | Blue component 9.64 | Total average 8.23 |
| Bilinear demosaicing RMS of reconstruction errors | Red component 13 | Green component 6.26 | Blue component 12.6 | Total average 11.05 |

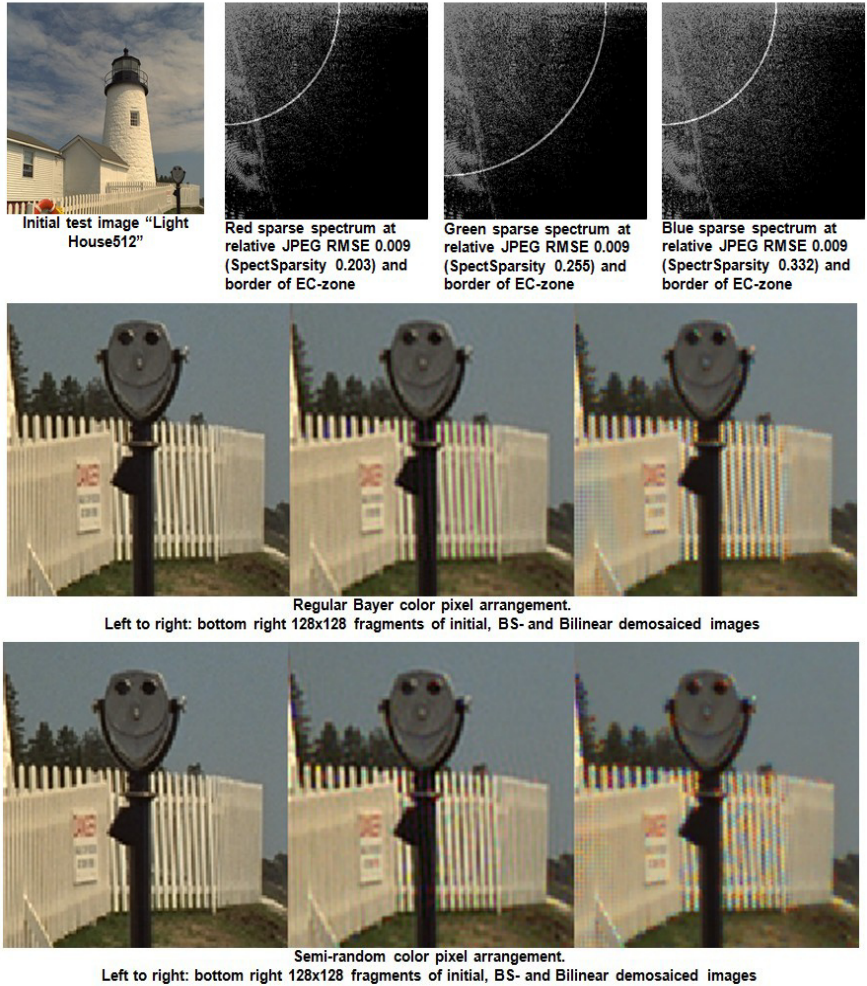

Figure 14. Comparison of BS-demosaicing and Bilinear demosaicing of a test color image with regular Bayer and semi-random pixel arrangements. Top row (left to right): test image of 512x512 pixels and sparse spectra of its red, green, and blue components (white dots) along with borders of the corresponding chosen EC-zones (white solid line). Middle and bottom rows (left to right): 128x128 pixel bottom right fragments of the initial test image and of BS- and Bilinear demosaiced images for "Regular Bayer" and "Semi-random" color pixel arrangements, correspondingly.

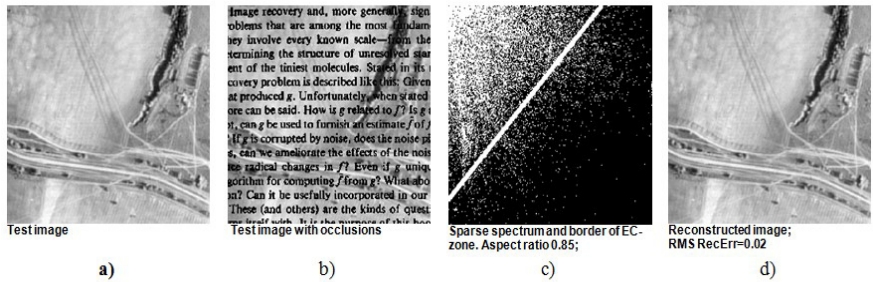

Figure 15. An example of using image BS-reconstruction for image in-painting: a) test image; b) test image occluded with a text; c) test image sparse DCT spectrum at JPEG RMSE 4.37 (white dots) and border of spectrum EC-zone used for image reconstruction; d) reconstructed test image (reconstruction RMSE 0.02 gray levels; peak signal to RMSE ratio 82.1 dB).

## Image in-Painting

Image in-painting is recovering images obstructed by foreign occlusions. In frequent cases image foreign occlusions can be reliably detected by means of one or another image segmentation method. In such cases, in-painting of image occlusions can be carried out with exactly the same algorithm as the above described iterative algorithm of image BS-reconstruction from sparse samples, in this case from those samples that are found not occluded. An illustrative example is presented in Figure 15.

In this example a test image (Figure 15, a) is occluded by a text (Figure 15, b), which is completely black and thanks to this its pixels can be easily detected by their gray level. Figure 15, d) presents the image BS-reconstructed from its samples not marked as occlusions. The image EC-zone used for the reconstruction is shown in Figure 15, c).

## Image Reconstruction from Their Sparsely Sampled or Decimated Projections

In computed tomography, it quite frequently happens that body slice occupies only a fraction of the image entire area. This means that slice projections are Radon transform "Bounded Spectrum" functions. Therefore, whatever number of projections or of their samples is available, a certain number of additional projections or samples, commensurable, according to the discrete sampling theorem ([14], [15]), with the size of the slice empty zone, can be obtained, and the corresponding resolution increase in the reconstructed images can be achieved using the method of image BS-reconstruction ([14]). This option is illustrated in Figure 16 and Figure 17.

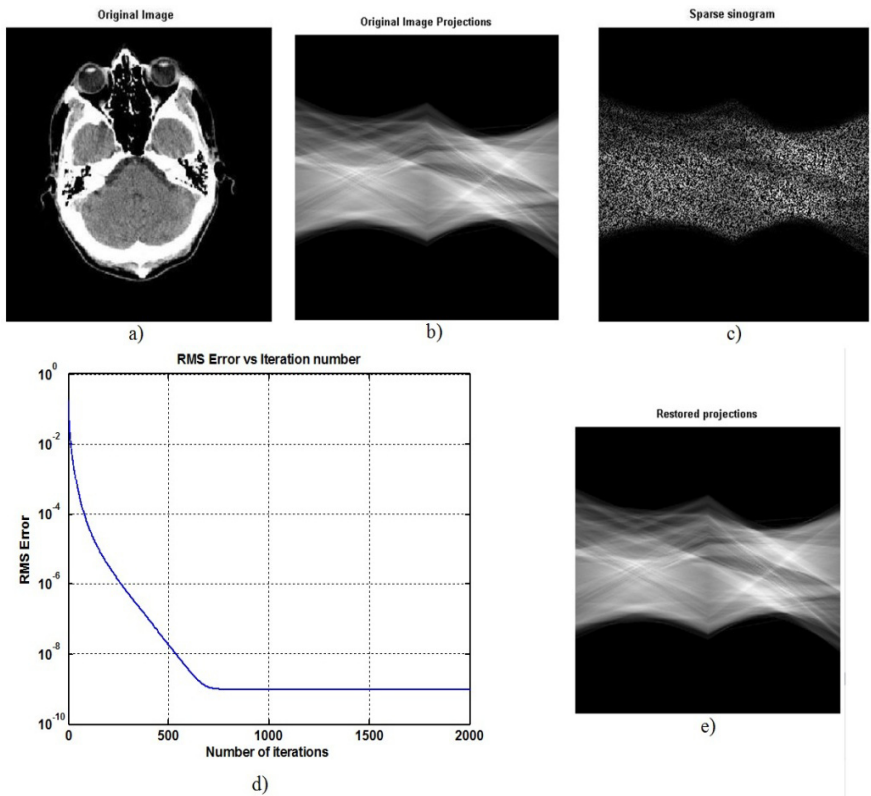

Figure 16. Recovery of randomly sampled slice projections: a) test image; b) its slice projections; c) slice projections randomly sub-sampled with rate 0.45; d) plot of slice projections reconstruction RMSE versus the iteration number; e) the recovered slice projections.

Figure 16 illustrates recovery of missing samples of an image slice projections sampled in random positions. In the experiment, it was found by means of simple segmentation of the test image shown in Figure 16, a) that the outer 55% of the image area is empty. Then the same percentage of projection samples selected randomly using the Matlab random number generator were zeroed (Figure 16, c). The rest of samples was used for recovering missing samples and, correspondingly, for image reconstruction by means of the iterative reconstruction algorithm identical to the above described image BS-reconstruction algorithm except that direct and inverse Discrete Cosine Transforms were replaced by direct and inverse Discrete Radon Transforms. At each iteration of the algorithm, the current set of projections is subjected to inverse Radon transform for obtaining a current estimate of the reconstructed image. Then the outer empty area of the reconstructed image is zeroed and this corrected image is subjected to direct Radon transform for obtaining a next estimate of slice projections. In the obtained projections, their available samples are restored and the process repeats. The plot of RMS of slice projections reconstruction errors versus the iteration number shown in Figure 16, d) and the result of recovering missing samples (Figure 16, e) show that virtually perfect recovery of missing 55% samples of slice projections is possible with the iterative reconstruction algorithm after few hundreds of iterations.

Figure 17 illustrates that recovery of completely missing projections is also possible. In the experiment, every second projection of the test image shown in Figure 16, a) was removed (Figure 17, b) and then all initial projections were recovered (Figure 17, c) by the above described iterative algorithm that makes use of the fact that the outer 55% part of the image area is known to be empty. This result allows suggesting that for such cases, when half or bigger part of the image area is known to be empty, one can use the method of image BS-reconstruction to achieve image reconstruction with super-resolution that corresponds to larger number of image projections than are available.

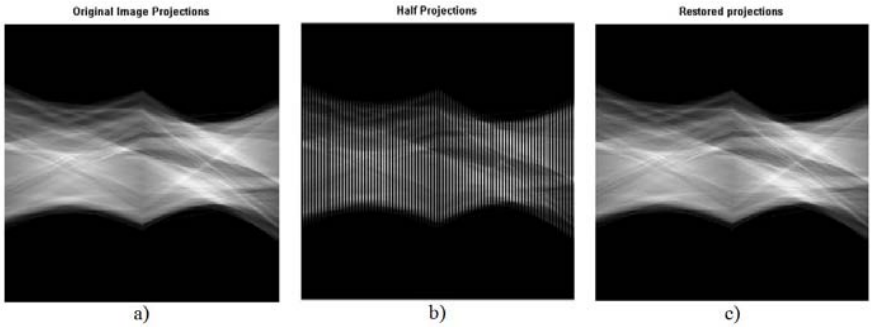

Figure 17. Recovery of missing image projections: a) original projections of the test image of Figure 16, a); b) decimated projections with every second projection removed; c) slice projections recovered from decimated slice projections (b) using the iterative reconstruction algorithm.

## Image Reconstruction from Their Sparsely Sampled Fourier Spectra

There exist some imaging devices (e.g., some healthcare scanners), where sampling is done in a transform domain. The proposed ASBSR-method can be used in such devices in those frequent cases, when it is known that object image is surrounded by some empty space. Figure 18 demonstrates this option on an example of image reconstruction from its sparsely sampled Fourier spectrum.

In this example, Fourier spectrum of a test image bounded by a circular binary image mask was randomly sampled with sampling rate equal to the ratio of the image bounding circle area to the area of the entire image frame. Additionally, the spectrum was bounded by a circular binary spectral mask with radius equal to the highest horizontal and vertical spatial frequency of the baseband. This gives an additional $1 - \pi/4 = 21.5\%$ saving in the number of spectrum samples.

For image reconstruction, an iterative algorithm was used. At each iteration, the iterated spectrum is inverse Fourier transformed for obtaining an iterated reconstructed image and then the latter is multiplied by the bounding circular image mask and Fourier transformed. Samples of the obtained spectrum in positions of available original ones are restored, and the spectrum is bounded by the circular binary spectral mask to form an iterated spectrum for the next iteration. As a spectrum zero order approximation, from which the iterative reconstruction starts, the initial sparsely sampled and bounded spectrum was used.

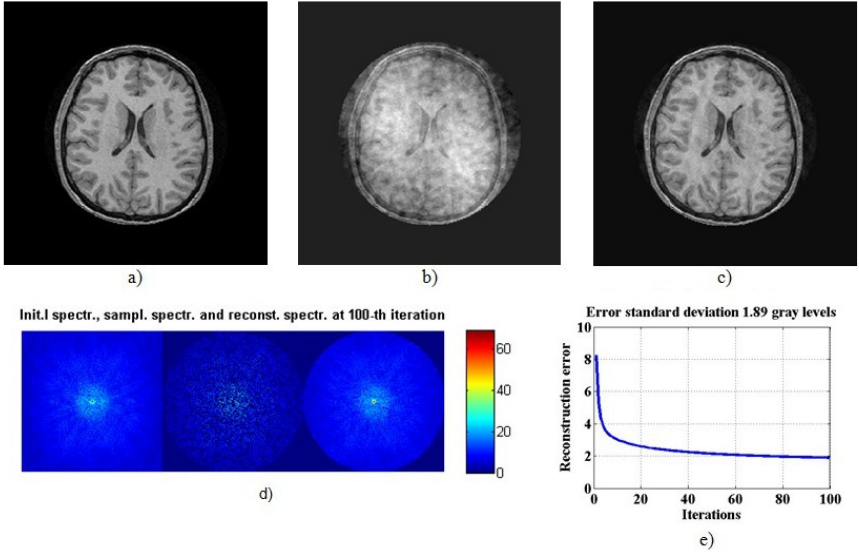

Figure 18. Image reconstruction from its sparsely sampled spectrum: a) test image bounded by a binary circular mask with radius equal to 0.35 of the image size; b) reconstructed image at the first iteration, in which the circular bounding mask can be seen; c) reconstructed image at the 100-th iteration (RMS reconstruction error is 1.89, peak signal to RMSE ratio is 49 dB); d) from left to right: Fourier spectrum of the test image, this spectrum randomly sampled with sampling rate $\pi \times 0.35^2 \times \pi/4 = 0.3$, and reconstructed spectrum (for display purposes, absolute values of spectral samples are displayed raised to power 0.3 and are shown color coded with Matlab color map "jet"); e) plot of reconstruction RMSE vs the number of iterations.

## Image Reconstruction from the Modulus of Its Fourier Spectrum

In order to enable image reconstruction from the modulus of its Fourier spectrum using the suggested ASBSR-method, object should be imaged through a randomized binary (opaque-transparent) mask that produces occlusions in the object image. Fraction of the transparent area of the mask should be equal to or larger than the required by the ASBSR-method sampling rate, i.e., than the fraction of the area occupied in the spectrum baseband by the chosen for this image EC-zone of the image DCT spectrum. Measured is the modulus of Fourier spectrum of the object occluded by the mask.

Reconstruction of the object image from the measured modulus of Fourier spectrum of its occluded copy is carried out in two stages. At the first stage, object image with occlusions is reconstructed from the modulus of its Fourier spectrum using an iterative Gerchberg-Papoulis type algorithm. Iterations reconstruct the image with occlusions and estimates of phase component of its Fourier spectrum. At each iteration, a current spectrum phase component estimate is combined with the measured spectrum modulus to form a complete spectrum estimate, which is inverse Fourier transformed to obtain a current estimate of the reconstructed image with occlusions. Then this image is multiplied by the binary occlusion mask, which restores the occlusions, and then Fourier transformed. Phase component of the obtained spectrum is used as the next estimate of the image phase component, and iterations are repeated. As a starting zero order estimate of the image spectrum phase component, phase component of the Fourier spectrum of the binary occlusion mask can be used.

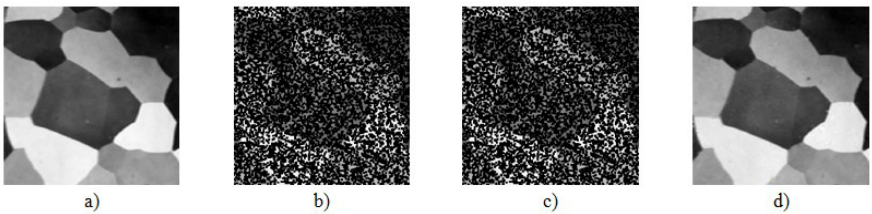

Figure 19. Image BS-reconstruction from the modulus of its Fourier spectrum: a) test image; b) test image occluded by randomly placed 3x3 pixel opaque squares; c) occluded image reconstructed from the modulus of its Fourier spectrum (reconstruction RMSE is $10^{-3}$; peak signal to RMSE ratio 107 dB); d) test image BS-reconstructed from the reconstructed occluded image (c) (reconstruction RMSE is 1.79; peak signal to RMSE ratio is 43.1 dB).

At the second stage of image reconstruction, the reconstructed image with occlusions is used for BS-reconstruction of the entire image in the same way as in the above described image in-painting. An illustrative example is shown in Figure 19.

## CONCLUSION

The problem of minimization of the number of measurements needed for digital image acquisition and reconstruction with a given accuracy is addressed. Specifically:

- It is shown that the theoretical lower bound of sampling rate sufficient for reconstructing an image of $N$ samples with a given RMSE is defined by image spectrum sparsity for this level of image sparse approximation error, i.e., by the ratio of the area of image spectrum Energy Compaction (EC-) zone formed from the largest spectral coefficients, which reconstruct the image with the given RMSE, to the area of the image baseband defined by its $N$ samples.
- Data saving capability of compressed sensing in image acquisition is disputed and it is shown that, for the range of sparsities of natural images from $10^{-1}$ to $10^{-3}$, the required number of measurements must exceed the theoretical minimum from 3 to 8 times, which implies that compressed sensing is far from reaching the theoretical minimum of signal sampling rate.
- A method of image Arbitrary Sampling and Bounded Spectrum Reconstruction (ASBSR-method) is described capable of reconstructing images sampled with rates that come near to the theoretical minimum. The method assumes representing images in a domain of one of transforms with sufficiently good energy compaction capability, approximating image spectra EC-zones by one of a few standard shapes, image sampling in arbitrary positions, and reconstructing images with spectrum bounded by the chosen shape. The method is insensitive to noise in sampled data and allows to secure obtaining a desired resolving power of reconstructed images.

- The workability of the method is demonstrated by results of its experimental verification using DCT as the image transform, image sampling over uniform sampling grids with random jitter, and an iterative Gerchberg-Papoulis type reconstruction algorithm. These results confirm that the sampling rate required by the method is equal to the theoretical minimum for the chosen EC zone shape, i.e., to the shape's relative area. This means that, for the class of images with the chosen EC-zone, the method is non-redundant. Sampling rates of particular images tested in the experiments exceeded their actual sparsities 1.5-1.8 times. This is the order of magnitude of the ultimate sampling redundancy of the method associated with approximating image spectra EC-zones by one of the standard shapes.

- Examples of application of the ASBSR-method for color image demosaicing, image in-painting, image reconstruction from sparsely sampled or decimated projections, image reconstruction from its sparsely sampled Fourier spectrum, and image reconstruction from the modulus of its Fourier spectrum demonstrate potential applicability of the method for solving other underdetermined inverse imaging problems.

RR